\pdfoutput=1 

\RequirePackage{fix-cm}
\documentclass{svjour3}                     
\smartqed  

\usepackage{graphicx}
\usepackage{multicol,footmisc,multirow,booktabs}
\usepackage[caption=false]{subfig}
\usepackage[acronym,toc]{glossaries}	
\usepackage[dvipsnames]{xcolor}					
\usepackage{amsmath}
\usepackage{graphicx}
\usepackage{verbatim}
\usepackage[normalem]{ulem}

\usepackage[
backend=biber,
sorting=none
]{biblatex}

\newacronym{ROS}{ROS}{Robotic Operating System}
\newacronym{EKF}{EKF}{Extended Kalman Filter}
\newacronym{IMU}{IMU}{Inertial  Measurement  Units}
\newacronym{VR}{VR}{Virtual Reality}
\newacronym{SVM}{SVM}{Support Vector Machines}
\newacronym{RSVM}{RSVM}{Ranking  Support Vector Machines}
\newacronym{DOF}{DOF}{Degrees of Freedom}
\newacronym{DNN}{DNN}{Deep Neural Network}
\newacronym{FNN}{FNN}{Feed-forward Neural Network}
\newacronym{DFNN}{DFNN}{Deep Feed-forward Neural Network}
\newacronym{RNN}{RNN}{Recurrent Neural Network}
\newacronym{PCA}{PCA}{Principal Component Analysis}
\newacronym{LSTM}{LSTM}{Long-Short Term Memory}
\newacronym{wrt}{wrt}{with respect to}

\begin{document}

\title{Mapping Surgeon’s hand/finger motion during conventional microsurgery to enhance Intuitive Surgical Robot Teleoperation
\thanks{This work was supported by the European Union’s Horizon 2020 research and innovation programme under grant agreement No 732515.}
}




\author{ Mohammad Fattahi Sani \and Raimondo Ascione \and Sanja Dogramadzi* 
}


\institute{* Corresponding author Sanja Dogramadzi, University of Sheffeild\\
              \email{s.dogramadzi@sheffield.ac.uk}           
}

\date{Received: date / Accepted: date}

\maketitle

\begin{abstract}
:\\
\textbf{Purpose:}
Recent developments in robotics and artificial intelligence (AI) have led to significant advances in healthcare technologies enhancing robot-assisted minimally invasive surgery (RAMIS) in some surgical specialties. However, current human-robot interfaces lack  intuitive teleoperation and cannot mimic surgeon’s hand/finger sensing and fine motion. These limitations make tele-operated robotic surgery not suitable for micro-surgery and difficult to learn for established surgeons. We report a pilot study showing an intuitive way of recording and mapping surgeon’s gross hand motion and the fine synergic motion during cardiac micro-surgery as a way to enhance future intuitive teleoperation.
\\
\textbf{Methods:}
We set to develop a prototype system able to train a Deep Neural Network (DNN) by mapping wrist, hand and surgical tool real-time data acquisition (RTDA) inputs during mock-up heart micro-surgery procedures. The trained network was used to estimate the tools poses from refined hand joint angles. Outputs of the network were surgical tool orientation and jaw angle acquired by an optical motion capture system.
\\
\textbf{Results:} Based on surgeon’s feedback during mock micro-surgery, the developed wearable system with light-weight sensors for motion tracking did not interfere with the surgery and instrument handling.  The wearable motion tracking system used 15 finger/thumb/wrist joint angle sensors to generate meaningful data-sets representing inputs of the DNN network with new hand joint angles added as necessary based on comparing the estimated tool poses against measured tool pose. The DNN architecture was optimized for the highest estimation accuracy and the ability to determine the tool pose with the least mean squared error. This novel approach showed that the surgical instrument ’s pose, an essential requirement for teleoperation, can be accurately estimated from recorded surgeon’s hand/finger movements with a mean squared error (MSE) less than 0.3\%
\\
\textbf{Conclusion:} We have developed a system to capture fine movements of the surgeon’s hand during micro-surgery that could enhance future remote teleoperation of similar surgical tools during micro-surgery. More work is needed to refine this approach and confirm its potential role in teleoperation.


\end{abstract}

\section{Introduction}
\label{intro}

Robot-Assisted Surgery (RAS) is preferred to conventional laparoscopic surgery in some areas due to reduced invasiveness, superior ergonomics, precision, dexterity and intuitive interaction  \cite{alemzadeh2016adverse}. 
resulting at times in shorter procedure and hospitalization times \cite{rodriguez2009robotics}. Intuitiveness of surgical tele-operation in RAS is essential in ensuring safety while obtaining the right level of procedural accuracy and effectiveness. Effective teleoperation depends on accurate mapping of the surgeon’s hand/fingers operating motion and the flawless translation of these fine movements from the surgeon’s master to the surgical instrument slave. 


Sub-optimal teleoperation has limited the widespread use of safe and effective RAS. For example, while RAS is widely utilized in urology, its adoption is limited or non-existent in micro-surgical specialties. Another limiting factor is that the surgeon’s master of current RAS systems is very different from conventional microsurgical instruments which makes it difficult for established surgeons to adopt RAS unless subjecting themselves to a new training. Moreover, the precision of slave positioning in e.g.  Da Vinci surgical robot is not sufficient for surgeries requiring higher precision. 

Cardiac surgery and other specialty areas have seen little RAS penetration and some safety concerns  \cite{quint2018role, alemzadeh2016adverse}. In cardiac surgery, for example, the limited access to the heart, no space available for the slave instrument maneuvering and close proximity to other vital structures require finer slave instrument movements and a superior teleoperation to that available in current robotic systems. For example, the Da Vinci master console uses a pair of handles  \cite{simorov2012review} to control the slave end-effector with 3 degrees of freedom (DOF), open/close grasper and 2 DOF in the wrist. While this appears to be effective for urology, for MIS tools that require higher complexity and dexterity, the master side of the system should be refined to support the feasibility, safety and efficacy of micro-surgery specialties through smoother and more intuitive tele-operation.

Proposed methods to teleoperate a surgical robot range from using handles at the surgeon’s master console \cite{simorov2012review} to a touch screen control \cite{mattos2014novel} supplemented with gaze, voice and foot pedals control \cite{mewes2017touchless}. Commonly used master stations in robotic surgery such as the da Vinci master station, Phantom Omni, haptic device neuroArm system \cite{sutherland2013evolution} were compared with novel approaches such as wireless or wearable data gloves and upper-body exoskeleton masters \cite{simorov2012review} suggesting  that future RAS technologies are more likely to use wearables as master device.

Hand/finger tracking has been investigated for applications ranging from teleoperation to motion analysis. To this end, inertial measurement units (IMUs), optical sensors, exoskeletons, magnetic sensing, and flex sensors-based systems have been used \cite{FattahiSani2019towards}. IMUs have also been used to complement or compensate errors of Kinect depth camera for skeletal tracking \cite{du2018imu}. However, depth camera is vulnerable to occlusions which prevent reliable pose estimation cite{li2019survey}. IMU sensors on thumb and index finger are sufficiently precise to authenticate in-air-handwriting signature using a support vector machine (SVM) classifier \cite{lu2017data}.







The concept of Robotic Learning from Demonstration (LfD) is attractive for teaching robot repetitive tasks from expert surgeons demonstrations as shown in \cite{reiley2010motion} for RAMIS or using surgical instruments’ trajectories to extract key features during complex surgical tasks  \cite{power2015cooperative}. Machine learning classification methods to evaluate surgeon’s skills by processing data extracted from the Da Vinci system have been shown in \cite{fard2016machine} or by tracking MIS tools in \cite{speidel2006tracking}. While in most studies surgeon’s movements have been used to analyze generated trajectories, the actual dexterity of surgeons’ hands/fingers during complex surgeries has not been studied.

Control of robotic minimally invasive surgical system using index finger and thumb gestures has been shown in \cite{itkowitz2013master} and \cite{itkowitz2015method}. However, using gestures to teleoperate a surgical robot is not intuitive or precise.   In our earlier  studies, we designed a hand tracking wearable system using IMU sensors placed on the hand digits and wrist to control a 4 DOF da Vinci Endowrist instrument.  \cite{Abeywardena2019Control}. We used the same, 12 DOF tracking wearable device to capture and analyse complex motion of surgeon’s hand during cardiac surgery procedures which helped us define the digits range of motion, workspace and the motion rate of change \cite{FattahiSani2019towards}. 

Here we show how surgeon’s hand and wrist motion during open access microsurgery is mapped to the surgical instrument motion with the aim to enhance a more effective, fine resolution, multi DOF tele-operation of surgical instruments for microsurgery to be used in future surgical robotic systems. To this end, our ultimate goal was to develop the first anthropomorphic prototype master concept based on learning the complexity of the fine synergic microsurgical motion

\section{Methodology}\label{sec:1}
\begin{figure}[htbp]
	\centerline{\includegraphics[scale=0.35]{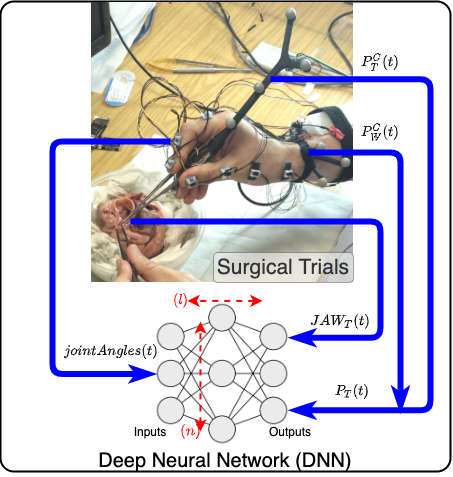}}
	\caption{Concept of the proposed idea: train a network during the open surgery and using that to map the hand finger motions to the surgical instrument. \\
	\textbf{Network outputs:} $P_T^C$ : tool pose in camera frame, $P_W^C$: wrist pose in the camera frame, $P_T$:tool pose with respect to the wrist, $JAW_T$: tool jaw angle (opening closing of the tool)\\ 
    \textbf{Network inputs: } joint angles in euler\\
    * $l$ and $n$ represent number of hidden layers and number of neurons in each hidden layer, respectively. }
	\label{fig:concept_ft}
\end{figure}

To develop the anthropomorphic prototype master concept based on mapping surgeon’s hands/fingers to the fine synergic motions of surgical instruments we disregard the gross motion of the robotic shaft holding the surgical instrument. The approach used is illustrated in  Fig~\ref{fig:concept_ft}.

A light weight wearable hand/finger/tool-status (open/close) custom made tracking system based on clinical feedback was designed to allow natural movements during surgery and capture complex movements generated during mock cardiac surgery scenarios. An optimized ANN architecture based on deep learning was built to estimate surgical tool position and orientation and validate the proposed hand-tool relationship and utilization of each hand/finger joint while performing the tool motion in a specific surgical task.

First, we developed motion tracking systems of the hand and the selected surgical tools - fine Castroviejo’s needle-holder and surgical forceps (needs reference) operated through the typical three-finger approach – thumb, index and middle finger. The tracking system captured two sets of data during typical mock cardiac surgery procedure performed by a senior heart surgeon. We derived a relationship between the two consecutive experiments conducted ex-vivo on animal samples using Castroviejo’s surgical tools. Data capturing focused on surgeon’s hand/wrist movements and concomitant Castroviejo tools’ poses. A trained neural network was used to map the Castroviejo motion using the hand joint angles. In addition, we used Gini metrics of decision tree regressor learning approach \cite{loh2014classification} to assess contribution of each finger and thumb joints in the performed surgical tasks.

\subsection{Hand and Tool Pose Tracking}

Surgical operations require complex dexterous manipulations of both hands so merely tracking the digit joint angles is not sufficient to fully represent hand poses and map them to the tool. After initial observations and analysis of cardiac open surgical procedures, the following key motion aspects were targeted to be captured:
\begin{enumerate}
    \item Surgeon's hand joint angles (digit and wrist joints)
    \item Global position and orientation of surgeon’s hands.
    \item Global position and orientation of the surgical instrument
    \item Surgical Instrument's Jaw angle (opening/closing (O/C) of the surgical instrument) 
\end{enumerate}

\begin{figure*}
\centering
	\subfloat[IMU sensors' attachment on the hand and joint names.]{
		\includegraphics[width=0.5\textwidth]{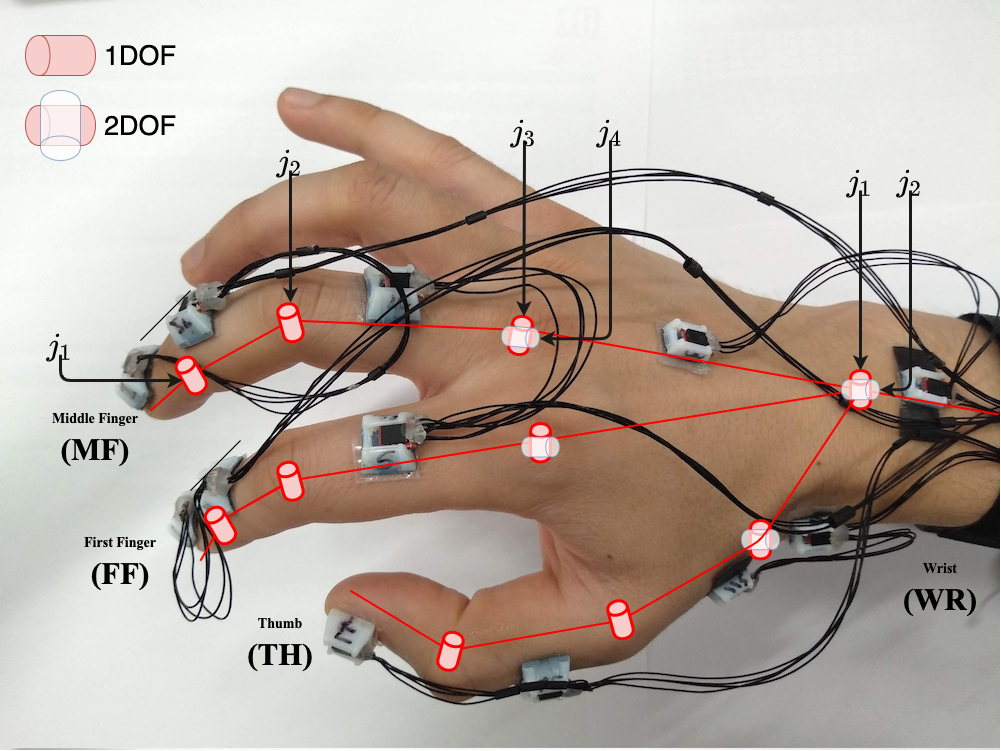}}
~
	\subfloat[\textbf{Top:}Custom data accusation board.\\ \textbf{Bottom:}Strain gauge attached on castroviejo tool.]{
	\includegraphics[width=0.4\textwidth]{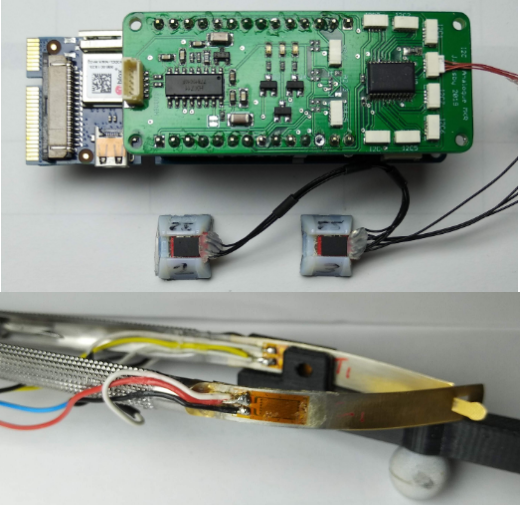}}
\caption{Sensors and the board}
\label{fig:sensors_and_the_board}
\end{figure*}

We used a chain of 12 IMUs to track hand and wrist joint angles (Fig.~\ref{fig:sensors_and_the_board} (a)). Hand digits are comprised of three joints, metacarpophalangeal (MCP), proximal interphalangeal (PIP), and distal interphalangeal (DIP) joint and represented in the hand kinematic model shown in our previous work \cite{FattahiSani2019towards}. Our hand/wrist tracking system is comparable with the latest commercial data gloves (e.g VR Manus) but features more sensing points in order to account for fine motion and a larger number of DOFs. Therefore, the index and middle fingers and the thumb, with the key roles in fine grasping and manipulation of Castroviejo instruments, are each tracked by 3 IMUs (Fig.~\ref{fig:sensors_and_the_board} (a)). 
The Castroviejo instrument is more complex for tracking due to its shape and size, particularly its open/close function for which we used four strain gauges attached to each side of the Castroviejo instrument (Fig.~\ref{fig:sensors_and_the_board} (b)) in a full Wheatstone bridge configuration.

\begin{figure}[htbp]
	\centerline{\includegraphics[scale=0.35]{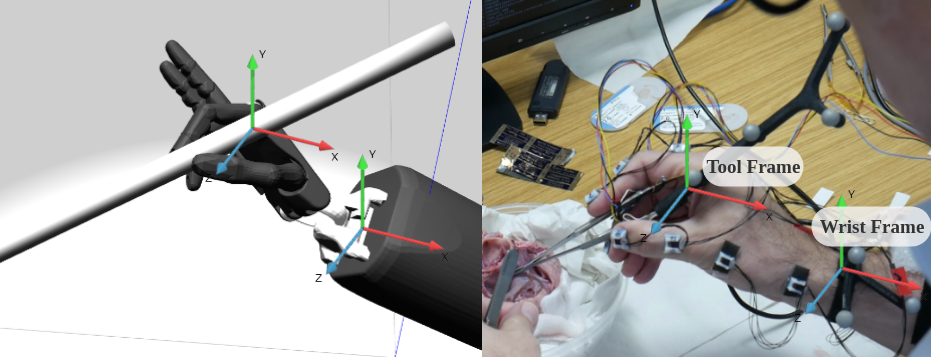}}
	\caption{\textbf{Left:} Hand pose and tool tracking using reflective markers, and \textbf{Right:} Gazebo simulation\cite{koenig2004design} of the acquired poses and orientations represented using the Shadow hand model\cite{ShadowHand}. }
	\label{fig:hand_tool_and_Shadow}
\end{figure}

The global position and orientation of surgeon’s hand and the surgical instrument are recorded using a Polaris Spectra optical sensor from Northern Digital Inc. (NDI) \cite{polaris2004northern} and a set of holders for reflective infrared markers designed and attached to the wrist and the Castroviejo tool (Fig.~\ref{fig:hand_tool_and_Shadow} (right)). Polaris Spectra tracks the markers in 6DOF and acquires the hand pose data using the proprietary ’NDI track’ software SawNDITracker library from Computer-Integrated Surgical Systems and Technology (CISST) \cite{deguet2008cisst} publishes data sent by NDI in the Polaris camera frame.

$P_T^C$ and $P_W^C$ are pose of the castro tool and the surgeon's wrist in the camera frame, respectively. We then calculate $P_T$ as a tool pose with respect to the wrist:

$P_T = [\phi_T(t), \theta_T(t), \psi_T(t)]$

where $\phi_T(t)$, $\theta_T(t)$, and $\psi_T(t)$ are roll, pitch and yaw angles of Castroveijho tool in space.

\begin{figure}[htbp]
	\centerline{\includegraphics[scale=0.5]{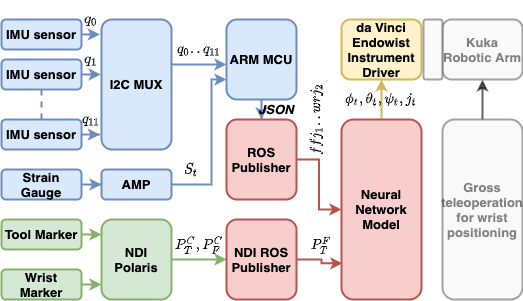}}
	\caption{Overall block diagram of the proposed algorithm. Inputs of the neural network is joint angles and tool status comming from strain gauge. outputs of the network is orientation of the surgical instrument and jaw opening and closing of the instrument.}
	\label{fig:block_diag}
\end{figure}

The system architecture is presented in Fig.~\ref{fig:block_diag}. The data are collected by: 1) IMU sensors, 2) strain gauges, and 3) Polaris motion capture system. We used a \gls{ROS} \cite{ros2018} on Ubuntu 16.04 for data recordings and processing. A da Vinci Endowrist tool was initially used to validate the hand/tool mapping architecture.

A custom data acquisition and data transmission system was designed to capture finger and thumb joint angles. Since middle and index fingers and thumb have the key role in fine grasping and manipulation, we track their poses using twelve inertia measurement units (BNO055 sensors from Bosch, see Fig.~\ref{fig:sensors_and_the_board} (a)). BNO055 chip has two different I2C bus address and we use PCA9548A 8-channel I2C switch to connect every pair of sensors to a single bus. Core of the system is a Microchip ATSAMD21 (Arm Cortex-M0+ processor) running on 48MHz, which is responsible for setting a correct configuration of the I2C switch, reading the orientation of each sensor, and sending the values to the computer through a USB port. All data captured by the board are framed as JSON data structures. The micro-controller also initializes all the sensors and puts them into calibration mode, if required. I2C bus is operated at 400KHZ frequency. IMU sensors produce quaternions and their relative values are calculated to obtain joint angles for each finger/thumb/wrist joint. These joint angles are then stored and visualized in Gazebo \cite{koenig2004design} and Rviz \cite{hershberger2019rviz} before used as inputs into the mapping model.

Assuming $q_0 - q_{11}$ are corresponding quaternion values from the 12 IMU sensors, a relative quaternion for each joint can be calculated. For instance $q_0^1$ represents $q_0$ with respect to $q_1$ can be found as follows:

\begin{equation} \label{eq1} 
q_0^1=q_1 \otimes q_0^{-1}
\end{equation} 
 where $q_0^{-1}$ is inverse quaternion and can be found by negating the w-component of quaternion : 
 \begin{equation} \label{eq2} 
q^{-1}=[q_x, q_y, q_z, -q_w]
\end{equation} 

Calculating the relative angles in quaternion will help us to avoid problems associated with euler angles. Final joint angles however are transformed to euler angles because of representation and matching with other signals.

\begin{equation} \label{input3} 
\begin{aligned}
X=[ff_{j1}, ff_{j2}, ff_{j3}, ff_{j4}, mf_{j1}, mf_{j2}, mf_{j3}, mf_{j4}, ...\\
th_{j1},th_{j2}, th_{j3}, th_{j4}, th_{j5}, wr_{j1}, wr_{j2}]
\end{aligned}
\end{equation} 



Data from the four strain gauges are sampled using an ADC (HX711, AVIA Semiconductor \cite{hx711201424}) with an on-chip active low noise programmable gain amplifier (PGA) with a selectable gain of 32, 64 and 128. $JAW_T(t) $ is jaw angle of castro tool  which is represented in euler. Assuming castro tool's jaw angle can vary between $30^{\circ}$ and $0^{\circ}$ in fully open and close scenarios, $JAW_T(t) $ can be calculated by mapping measurements from strain gauges.

The main MCU synchronises data from the strain gauges and the IMU and sends them to the PC at the rate of 50Hz through the USB.

\subsection{Data collection}

The motion tracking data were collected from an experienced cardiac surgeon performing typical cardiac tasks (coronary and mitral valve repair surgery) on an ex-vivo animal heart, which included all the different instances of cutting, suturing and knotting. 

\begin{figure*}
\centering
	\subfloat[Data collection during the heart surgery on ex-vivo animal heart.]{
		\includegraphics[width=0.5\textwidth]{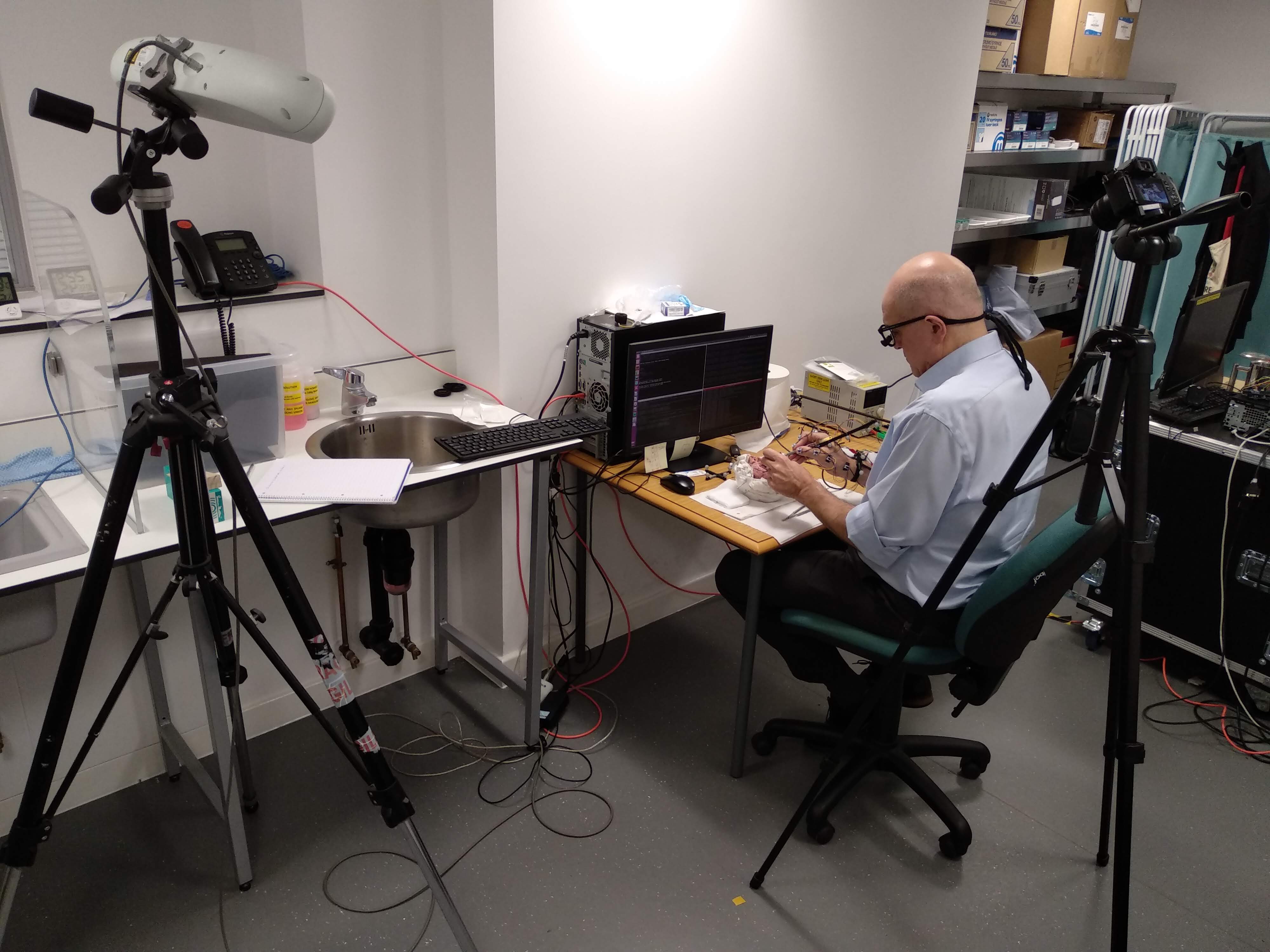}}
~
	\subfloat[Data collection during typical open heart surgery tasks on an ex-vivo porcine heart.]{
	\includegraphics[width=0.5\textwidth]{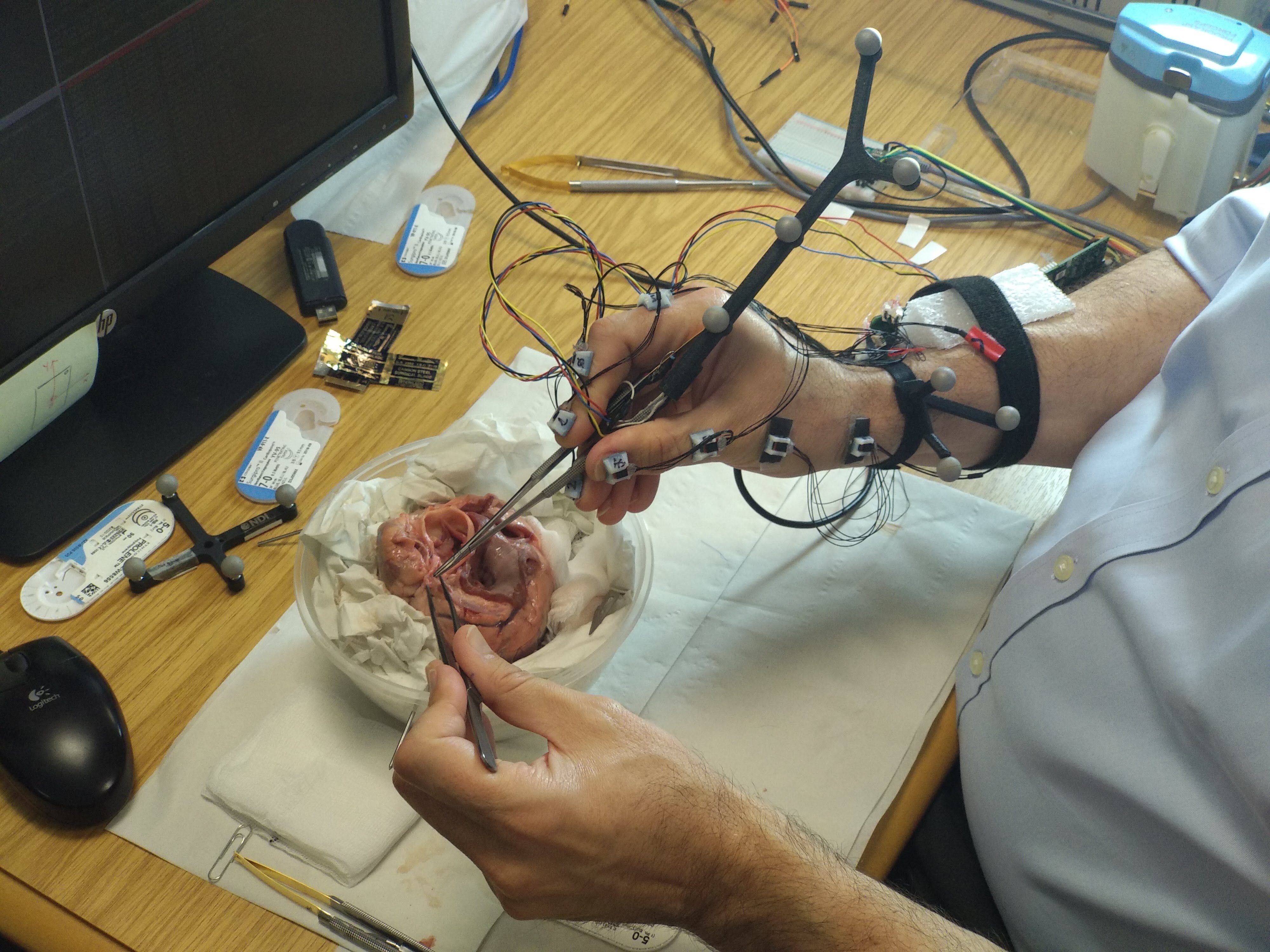}}
\caption{Data collection}
\label{fig:data_collection}
\end{figure*}

\begin{figure}[htbp]
	\centerline{\includegraphics[scale=0.31]{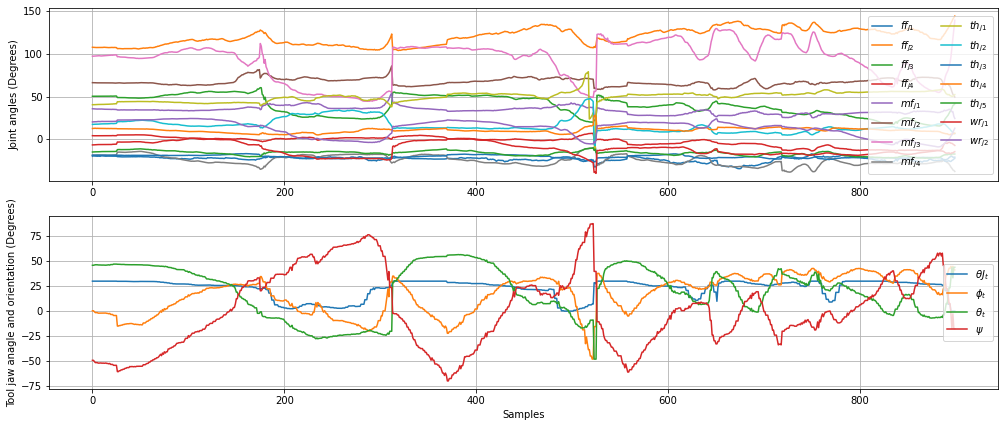}}
	\caption{sample data collected during the surgery showing joints angles as input and tool orientation and tool jaw angle as output}
	\label{fig:Hand-Tool-da-Vinci}
\end{figure}

The obtained data sets were pre-processed to filter out incomplete or missing data e.g occlusions of the tool or wrist markers that occasionally caused data loss. In the next stage input and output values are normalized to get better performance from neural network. All 10000 samples at the sample rate of 30Hz in the dataset are timestamped values of hand joint angles, tool orientation, and tool jaw angle. The mock surgical experimental setup for the data collection is shown in Fig~\ref{fig:data_collection} with a representative graph of the collected data shown in Fig~\ref{fig:Hand-Tool-da-Vinci}.

\subsection{Hand-Tool modeling}
Due to the non-linear multi input-multi output nature of the collected data, simple machine learning methods such as multivariate linear regression, or Support Vector Regression (SVR) do not provide satisfactory results. A \gls{DNN}, however, can be efficiently trained to map multi DOF tool motion from the corresponding hand motion. Here we use two types of neural networks, Deep Feed-forward neural network (DFNN) and a Long-Short Term Memory (LSTM) neural networks, and compare their performance in terms of time and accuracy. \gls{DFNN} is a FNN network that has more than two layers, which enables the network to learn more complex patterns of data. A \gls{LSTM} neural network is a kind of \gls{RNN} that has feedback connections between the layers which enables learning a sequence of data without facing limitations of the RNNs such as the ‘vanishing gradient’ \cite{da2019using, gers1999learning}.

 A \gls{DNN} using Keras library in Python with $n$ number of neurons in each hidden layer and $l$ number of hidden layers is used here. We use both \gls{DFNN} and \gls{LSTM} neural networks and compare their performance in terms of time and accuracy.

Inputs of the neural network at a time of $t$ is comprised of 15 hand joint angle inputs in Euler representation:
\begin{equation} \label{input} 
\begin{aligned}
X(t)=[ff_{j1}, ff_{j2}, ff_{j3}, ff_{j4}, mf_{j1}, mf_{j2}, mf_{j3}, mf_{j4}, ...\\
th_{j1},th_{j2}, th_{j3}, th_{j4}, th_{j5}, wr_{j1}, wr_{j2}]
\end{aligned}
\end{equation} 
(see Fig.~\ref{fig:sensors_and_the_board} (a) for notation) and output of the network during the training is comprised of:
\begin{equation} \label{input2} 
Y(t)=[P_T(t), JAW_T(t)] = [\phi_T(t), \theta_T(t), \psi_T(t), JAW_T(t) ]
\end{equation} 
where $P_T$ is surgical tool pose and $JAW_t$ is surgical jaw angle.

After the neural network is adequately trained, the tracking markers and strain gauges attached on the tool are not required. The surgeon controls the tool by wearing the sensor chain while the neural network determines the outputs only from $JointAngles(t)$. The outputs of the network are $ [\phi_d(t), \theta_d(t), \psi_d(t),j_d(t)] $. This is one of the benefits of the proposed method as using vision based tracking methods like Polaris would not work when the markers are covered, and would need a more complicated setup

Details of selected network configuration is as follows:
\begin{itemize}
    \item Optimization Method: Adam with default parameters (learning rate ($lr$)$=0.001$, $\beta_1 = 0.9$, $\beta_2 = 0.999$, $decay=0.0$)
    \item Input neurons:15 features (joint angles) 
    \item output neurons: 4 ( 3 neuron for tool orientation + 1 for jaw angle )
    \item dataset size: 10000 samples
    \item train test split : 80\% for training, 20\% for testing
    \item number of epoches: 200
    \item loss function: mse
    \item FNN activation function: relu
\end{itemize}

The rationale of these parameters was to achieve the highest precision in estimating the tool orientation with a smallest and therefore fastest network size.




\section{Results and Discussion}

 \subsection{Network Complexity and Accuracy}
 
\begin{figure*}
	\centerline{\includegraphics[scale=0.40]{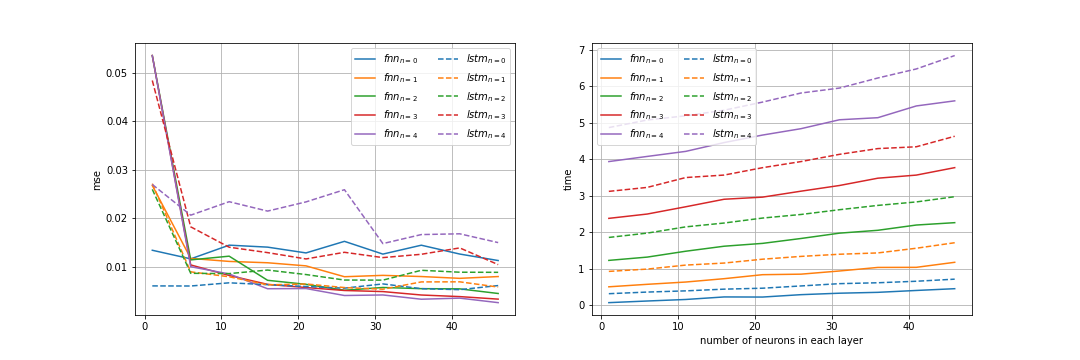}}
	\caption{Effect of number of neurons in each layer (n) and number of hidden layers (l) on estimation accuracy (mean squared error with normalized units) and execution time.}
	\label{fig:results}
\end{figure*}

\begin{figure}[htbp]
	\centerline{\includegraphics[scale=0.34]{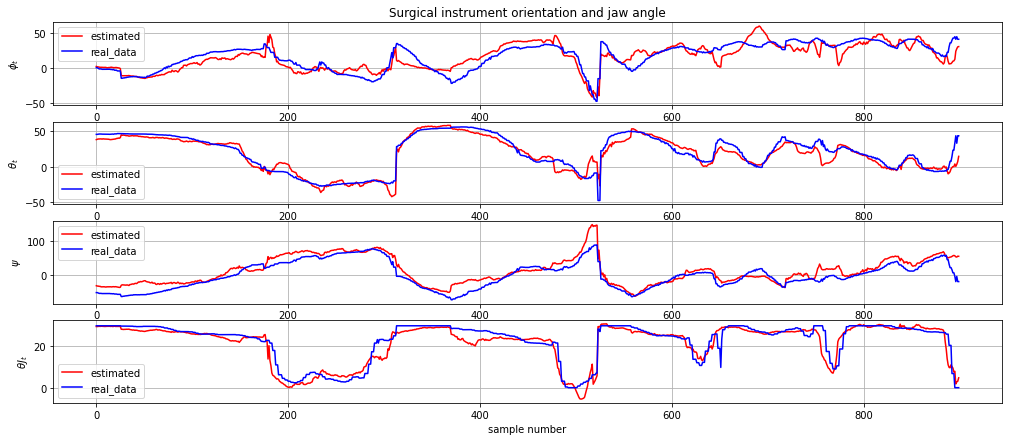}}
	\caption{Neural Network estimated values against unused data.\\ 
	$[\phi_T(t), \theta_T(t), \psi_T(t), JAW_T(t) ]$.(same slice as in Fig~\ref{fig:Hand-Tool-da-Vinci}) }
	\label{fig:tool_status}
\end{figure}

 In Fig.~\ref{fig:results} we report mean squared error of estimated tool pose compared to the ground truth obtained using Polaris sensor for different network depth and layer size for both DFNN and LSTM architectures. As it can be seen, having more than two deep layers increases the network complexity and affects the processing time. However, it appears that this does not reduce the error further. In addition, this suggests that increasing the number of the neurons over 20 in each layer does not improve network’s efficiency.  
Data shown in Fig~\ref{fig:tool_status} demonstrate that the network can accurately estimate tool orientation and jaw angle while the results in Table~\ref{tab:1} show an advantage of the LSTM neural network in terms of accuracy accompanied with almost three times longer processing time.

\begin{table}[ht]
\centering
\caption{Final results from \gls{LSTM} and \gls{DFNN} models }
\begin{tabular}{|c|c|c|c|c|c|c|c|c|}
\hline
\textbf{Model} & \multicolumn{4}{c|}{\textbf{DFNN}} & \multicolumn{4}{c|}{\textbf{LSTM}}\\ \hline
\textbf{joints} & \textbf{$\phi_T$} & \textbf{$\theta_T$} & \textbf{$\psi_T$}  & \textbf{$JAW_T$} 
& \textbf{$\phi_T$} & \textbf{$\theta_T$} & \textbf{$\psi_T$}  & \textbf{$JAW_T$} \\ \hline
\textbf{Root Mean Squared Error} & $7.5^{\circ}$& $7.0^{\circ}$& $12.1^{\circ}$&$3.1^{\circ}$
& $7.2^{\circ}$ & $6.1^{\circ}$ & $10.9^{\circ}$ & $1.8^{\circ}$ \\ \hline
\textbf{r2 score on training set} & \multicolumn{4}{c|}{91 \%} & \multicolumn{4}{c|}{92 \%}  \\ \hline
\textbf{r2 score on testing set} & \multicolumn{4}{c|}{90 \%} & \multicolumn{4}{c|}{91 \%}  \\ \hline
\textbf{Prediction time} & \multicolumn{4}{c|}{0.16 s} & \multicolumn{4}{c|}{ 0.46 s}  \\ \hline

\end{tabular}

\label{tab:1}
\end{table}

\subsection{Feature Importance}

\begin{figure*}
	\centerline{\includegraphics[scale=0.45]{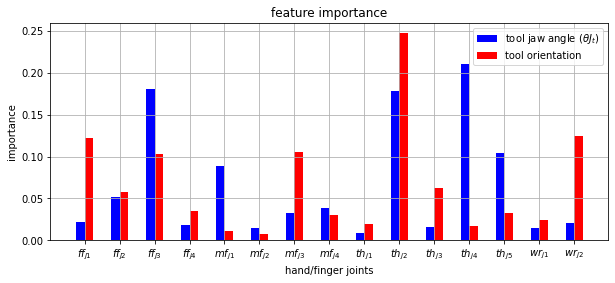}}
	\caption{Input feature importance with respect to corresponding outputs (tool orientation or jaw angle).}
	\label{fig:feature_importance}
\end{figure*}

\begin{figure*}
	\centerline{\includegraphics[scale=0.40]{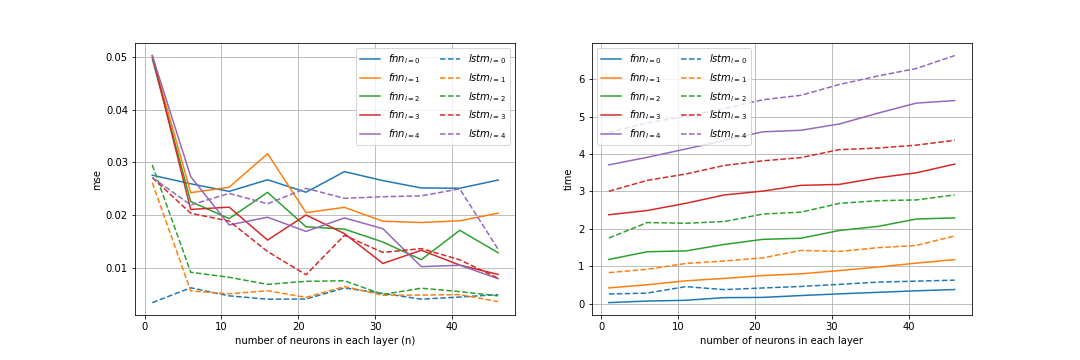}}
	\caption{result of the neural network trained with reduced inputs ( only 5 most important joints involved in surgery.(mean squared error with normalized units)}
	\label{fig:results_fi}
\end{figure*}

Feature importance identifies which joints are main contributors in creating certain types of tool motion. Decision Tree Regressor Gini importance or Mean Decrease in Impurity (MDI) \cite{breiman2001random} was used to establish a correlation between the hand joints and the tool variables. The importance of the input features for each output variable  is shown in Fig.~\ref{fig:feature_importance}. 
It can be seen that $th_{j2}$, $th_{j4}$, and $ff_{j3}$ are most important features for jaw opening of the tool while the tool orientation largely dependence on the $th_{j2}$, $wr_{j2}$, and $ff_{j1}$. This analysis also highlighted the less important joints such as $mf_{j2}$ which impact on teleoperation of the surgical instrument can be mostly neglected. 
When the two networks were trained and tested using only five most important joints, LSTM network demonstrated a better performance compared to DFNN as shown in Fig.~\ref{fig:results_fi}. These findings can be particularly useful in designing a new system with a reduced number of tracked joints and interfacing hardware complexity.

\subsection{Principal Component Analysis}

\begin{figure*}
	\centerline{\includegraphics[scale=0.40]{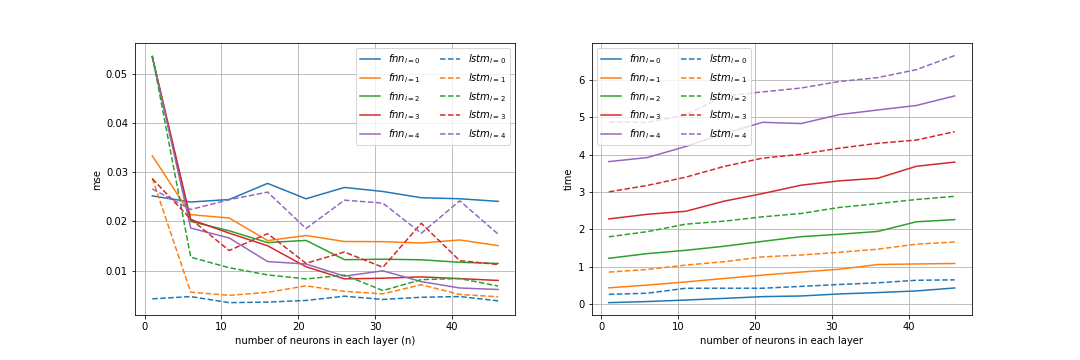}}
	\caption{result of the neural network trained with PCA ( 5 principal components as an input) (mean squared error with normalized units).}
	\label{fig:results_pca}
\end{figure*}

We also used Principal Component Analysis (PCA) to extract five Principal Components of the dataset to train the two networks (Fig.~\ref{fig:results_pca}). The LSTM neural network demonstrated slightly better performance in estimating the tool orientation.  LSTM networks are more complex than their DFNN counterparts due to the recurrent layers which consume more processing resources but they usually have better performance when it comes to time-series. The performed dimensionality reduction lowers the network complexity while maintaining the output accuracy, especially in LSTM networks. This is essential in teleoperation where latency plays a crucial role in the stability of teleoperation systems, specifically bi-directional teleoperation systems that require fast loop update rates.

\section{Conclusion and future work}

In this paper, we tested the hypothesis that motion of surgical instruments that perform fine movements in cardio surgery can be mapped to the hand/wrist motion of the surgeon and can potentially yield effective tele-operation in this type of surgeries. A wearable non-obstructive hand/wrist and surgical tool tracking system that can collect accurate multi-point data were built and tested. We created a mock up cardiac surgery test-bed where an experienced cardio surgeon performed typical surgical tasks. The collected datasets of hand/wrist joints as inputs and tool motion as outputs were used to train two types of deep neural networks - LSTM and DFNN.  We compared performance of the two networks using all captured inputs but also identified importance of each hand joint on the tool motion. Performance of the two networks was again compared using the smaller number of salient inputs to reduce the network complexity. The implemented optimization allows the use of fewer joint angles as control inputs in order to achieve the same output performance. This means that the surgeon could tele-operate a Castroviejo-like instrument using the same type of hand movements as in open access surgery, potentially decreasing/eliminating surgeons’ cognitive and muscular fatigue currently experienced with tele-operated surgical robots. Developing this approach further can address the need for surgical retraining to undertake same procedures through a completely different set of movements required in surgical robot tele-operation. Tele-operating a surgical robot using a wearable hand tracking system, combined with a VR head mounted, would as presented in \cite{danioni2020study} provide the surgeon with a unique ability to operate the robot closer to the patient and the operating theatre team or from any remote site while interacting with the team in the surgical theatre \cite{simorov2012review}.

Future advances in implementing the wearable tracking to a wider range of fine motion surgical instruments for cardiac and vascular surgeries would help establish and test the wearable tele-operation control concept. We have begun extending this framework to other surgical areas like arthroscopy with the similarly sized operating fields. While these results are encouraging, more relevant data are necessary to implement machine learning techniques which focuses us on conducting further user studies in this and other surgical scenarios.




%

\section*{Conflict of interest}

The authors declare that they have no conflict of interest.
\textbf{Funding:} This study was  supported  by  the  European  Union’s  Horizon  2020  research  and  innovation programme under grant agreement No 732515.

\section*{Compliance with Ethical Standards}
\textbf{Ethical approval :} All procedures performed in studies involving human participants were in accordance with the ethical standards of NHS Health Research Authority, REC Ref. 19/HRA/4102.

\textbf{Informed consent:} Informed consent was obtained from all individual participants included in the study.


\printbibliography

%
%

\newpage

\end{document}